# Abstract

# AI BASED FONT PAIR SUGGESTION MODELLING FOR GRAPHIC DESIGN

**Aryan Singh,** Microsoft Designer, singharyan@microsoft.com
**Sumithra Bhakthavatsalam,** Microsoft Designer, sumithrab@microsoft.com

One of the key challenges of AI generated designs in Microsoft Designer is selecting the most contextually relevant and novel fonts for the design suggestions. Previous efforts involved manually mapping design intent to fonts. Though this was high quality, this method does not scale for a large number of fonts (3000+) and numerous user intents for graphic design. In this work we create font visual embeddings, a font stroke width algorithm, a font category to font mapping dataset, an LLM-based category utilization description and a lightweight, low latency knowledge-distilled mini language model (Mini LM V2) to recommend multiple pairs of contextual heading and subheading fonts for beautiful and intuitive designs. We also utilize a weighted scoring mechanism, nearest neighbor approach and stratified sampling to rank the font pairs and bring novelty to the predictions.

## 1. Background

One of the key challenges of AI generated designs in Microsoft Designer is selecting the most contextually relevant and novel fonts for the design suggestions. The previous efforts of manually mapping design intent to fonts, even with great quality, do not scale very well for 3000 + fonts and numerous possible user intents when creating a design.

In this work, we create font visual embeddings, a font stroke width algorithm, a font category-to-font mapping dataset, an LLM-based category utilization description, and a lightweight, low-latency, knowledge-distilled mini language model (Mini LM V2). These components are used to recommend multiple pairs of contextual heading and subheading fonts for beautiful and intuitive designs. We also utilize a weighted scoring mechanism, nearest neighbor approach and stratified sampling to rank the font pairs and bring novelty to the predictions.

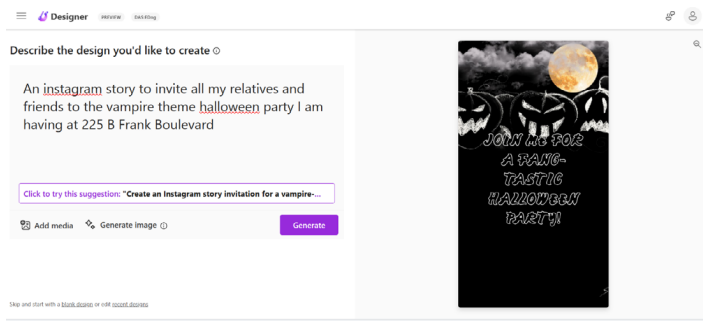

**Figure 1:** Mapping Halloween theme to contextual font.

AI driven contextual fonts lend much needed variety to AI generated designs which otherwise might surface repetitive and bland fonts. It helps users to express the key message creatively with an eye-catching typography.

## 2. Related work

There have been some previous efforts in representing fonts as visual embeddings for instance by Wang et al [1] but these are mostly convolution network based and are mainly applied for font recognition rather than recommendation. Further there have been studies to identify trends in typography design and attributes by Shinahara et al [2]. Some other studies have been done to associate personality traits and fonts like the ones by O'Donovan et al. [3], Brumberger et al. [4], Juni and Gross [5]. Another recent work in associating verbal context to fonts was done by Shirani et al. [6]. Most of these scale to a limited set of fonts and only work on recommending one relevant font for a design rather than recommending multiple relevant font pairs which are pursued in this work.

## 3. Methodology

Making design decisions is very hard because there is a lot of subjectivity. A good font can make design interesting and engaging. At the same time a bad font can break a good design. Hence selecting contextually relevant fonts is a very difficult problem. Also, most of the existing works suggest a single font for the design but do not suggest a pair of fonts for heading and subheading.

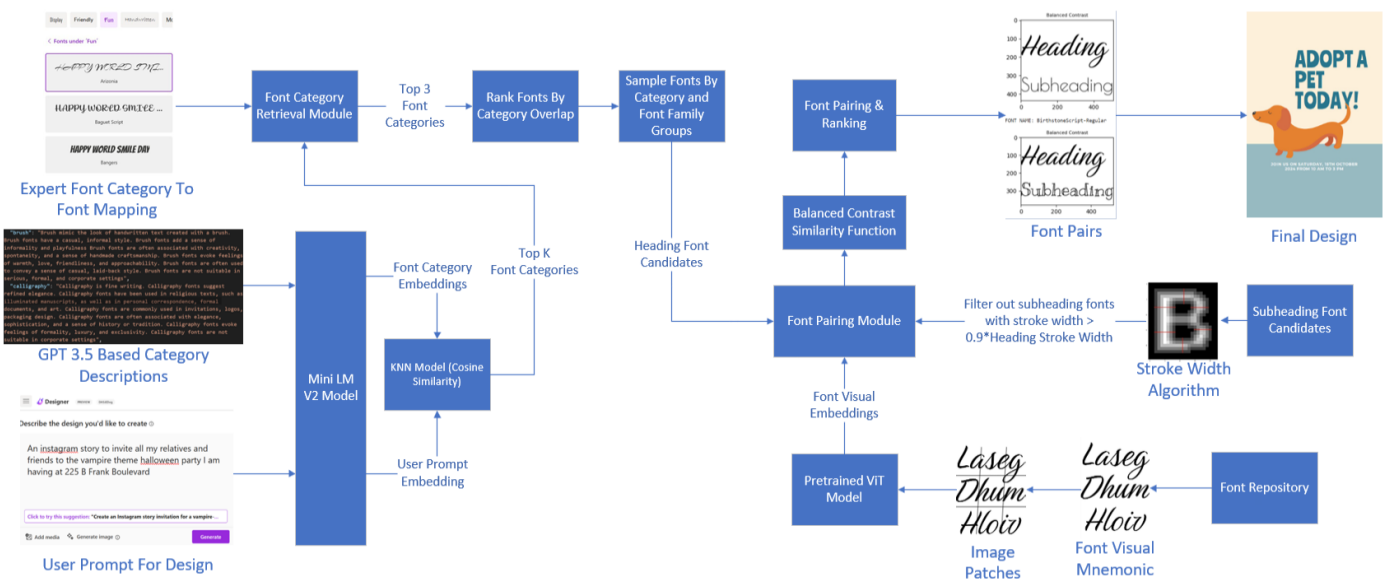

**Figure 2:** End to end details for the system.

Following are the steps for this approach, (1) The visual representation of fonts and the embedding representation of the font category descriptions are created. (2) The user prompt for design is represented as an embedding in the same state-space as the font category embeddings. (3) The top-n font categories are retrieved using KNN with cosine similarity. (4) The overlap of these font categories is used to rank fonts for heading and then the final heading fonts are sampled from ranked fonts, respecting the novelty of font families. Visual embeddings of these heading font candidates and all other fonts are extracted using the vision transformer large model. (5) The similarity and contrast scores between the visual embedding of heading font and subheading visual embedding candidates are extracted using the custom balanced-contrast distance function to determine better font pairs. (6) The stroke width algorithm is used to make sure the stroke width of subheading is less than heading font. This entire system is explained in Figure 2, and the entire pipeline is explained below.

**Expert defined font categories:** Font categories are devised by the typography expert, these are a combination of moods represented by font, appearance, and other attributes. A few of the categories are fun, cute, friendly, elegant, modern etc.

**Experts tag fonts to category:** Then around 3000 fonts are tagged with these categories. Some of the fonts are tagged into two categories, for example a font could be modern and curly at the same time.

**Font category description:** We used GPT 3.5 turbo prompt to generate usage details, history, mood, and design preferences for each of the font categories. Following is the GPT prompt used to generate these details:

> *What are [font category name] typefaces or [font category name] fonts category ? Where are they commonly used today, or used in the past? What emotions do they evoke? In which context are [font category name] category fonts suitable for usage, and in which context [font category name] category fonts should be avoided? Give your response in a crisp manner that's suitable for creating an embedding of your response using a large language model.*

Later, in consultation of typography experts and designers we modified these category descriptions a little bit to make them less noisy and more relevant for design usage.

**Font category embeddings:** We have a pre-trained Mini LM V2 6-layer model for sentence similarity. We pass the category descriptions through this model to find and store embeddings.

**User prompt embeddings:** The user enters a prompt to generate the design. We pass this prompt through the Mini LM V2 model to generate an embedding at run time.

**Top 3 categories:** Top 3 categories are retrieved by using a KNN approach with k as 3 and distance metric as cosine similarity. We get the top 3 categories such as fun, friendly, and cute.

**Retrieve and rank heading fonts by category overlap:**

> **User prompt:** A design for wedding invitation
> **Retrieved font categories:** Wedding, Cursive, Elegant
> **Mapped fonts:**
> - Wedding_Cursive: Font A, Font B
> - Wedding_Elegant: Font C, Font D
> - Wedding_Cursive_Elegant: Font E, Font F
> - Wedding: Font G
>
> **Font ranking:** Font E, Font F, Font A, Font B, Font C, Font D, Font G
> **Novelty:** Sample from each category
> - Wedding_Cursive_Elegant: Font E
> - Wedding_Cursive: Font A
> - Wedding_Elegant: Font C
> - Wedding: Font G

**Font visual embeddings:** Font visual embeddings are obtained by inferring font mnemonic images consisting of mnemonic text 'Laseg Dhum Hloiv' on the ViT large model [7]. This mnemonic text is fed into OpenCV's FreeType and putText functions to render the mnemonic images for each font. These images are then fed into a ViT large model to generate the visual embeddings for each font as shown below in Figure 3.

**Stroke width algorithm [8]:** We use font visual representation for font size 50 across all fonts and run distance transform and skeletonize operations to determine the stroke width of every font in the repository.

**Subheading fonts via font pairing algorithm:** Subheading fonts are selected via our font pairing algorithm which introduces a balanced contrast-based distance metric between visual embeddings of fonts to discover font pairs.

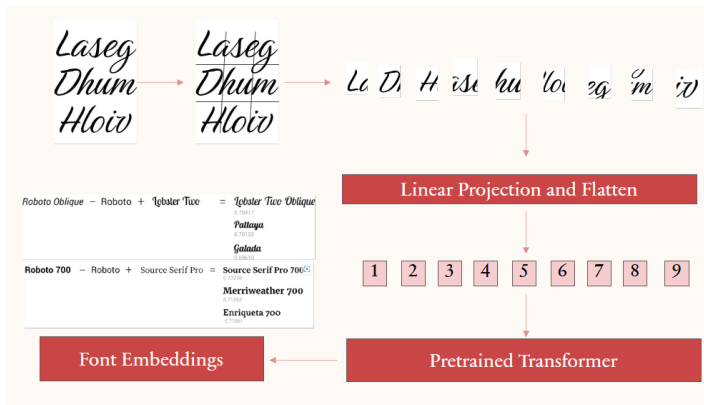

**Figure 3:** Vision Transformer for creating visual embeddings.

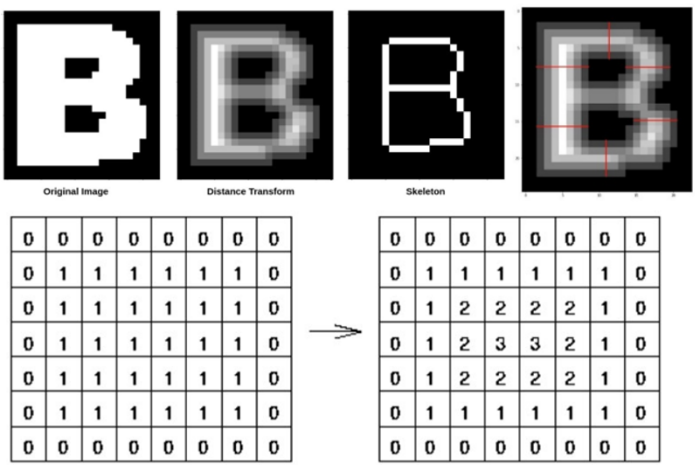

**Figure 4:** Stroke width algorithm for fonts. The binary visual image of font alphabet is first converted to a distance transform version to determine pixel wise thickness. Then skeletonization of the input image is done to find the central pixel. Stroke width is determined by multiplying the distance between center pixels and edges by 2.

## 4. Results

### 4.1 Font category retrieval evaluation

Firstly, we did an objective evaluation of different NLU models on a MTEB S2S [9] similarity benchmark. This benchmark consists of different data sets such as STS12, STS13 etc. which compare different natural language representation models on sentence-to-sentence similarity. The results of this benchmark data can be seen in Table 1.

Apart from this, font category retrieval was also evaluated on our custom prompt data collected via crowd sourcing within the team and evaluation was done by internal judges. Judges consisted of a team of **3 product managers** in the designer team who had worked on the typography features of Microsoft Designer. It was important that the judges were aware, both visually and conceptually, of the font categories and their constituent fonts. The labelers gave a relevance score on a grade of **1-5 to each of the top 3 font categories**, retrieved by the Mini LM model for each prompt. To score the output of each prompt, the average score of each font category was taken. Finally, we took an average across each of the **141 data points** to arrive at the final relevance score. Then different models were compared based on their average accuracy across all 141 prompts. This result can be seen in the **subjective evaluation column** in Table 1 below. The results of the font category from all these models can be found **here.**

Amongst the embedding based models, all of them apart from Simcse had comparable numbers on the subjective analysis. Considering the trade-off between latency and the subjective analysis results we went for the low latency (sub 100 milliseconds) Minilm V2 6-layer model, since our designs required near real time font suggestions.

The reasons for not going with the GPT 3.5 turbo model were twofold. Firstly, it had a higher latency of around 2 seconds which was around 20 times more than the Minilm V2 model. Secondly, due to the larger cardinality of font categories, GPT turbo tended to focus on just a few of them and sometimes hallucinated. All the evaluation results can be found in Table 1 above.

| |
|---|
| **Algorithm 1:** Font Pairing Algorithm |
| **Input:** Font Visual Embeddings, Fonts Stroke Width, Heading Font Candidates |
| 1. Given the current selected heading font is f and there are k different subheading font candidates, then m out of those k candidates are selected on basis of rule: |
| Stroke Width (Subheading Font j) <= 0.8*Stroke Width (Heading Font f) |
| We arrived at the coefficient of 0.8 empirically after running multiple experiments and visualizing the contrast of stroke width amongst the heading and subheading font candidates. |
| 2. Given that A is the visual embedding of font selected for heading and B is the visual embedding for the subheading font candidate. Here P is the positive coefficient of similarity obtained by Hadamard product of the same sign components in the heading and subheading font vectors. Similarly, N is the coefficient obtained by Hadamard product of opposite sign components in the heading and subheading font vectors. Then for $j^{th}$ subheading candidate: |
| $$P_j = \sum_{i=1}^{n} A_i B_i \ (where\ A_i B_i > 0)$$ |
| $N_j = -\sum_{i=1}^{n} A_i B_i\ (where\ A_i B_i < 0)$ |
| 4. Suppose there are m font candidates for subheading for the given heading font, we do min max normalization of positive and negative scores for all m subheading candidates. Suppose P is the tensor of positive scores across all m candidates and N is the tensor of negative scores: |

$$P_{norm} = MinMaxNorm(P)$$

$$N_{norm} = MinMaxNorm(N)$$

5. Out of the m candidates, the candidates which have a balance of negative and positive traits in terms of visual similarity with selected font are retained. This is done by:

$$P_{46} = \{P_{norm}\ if\ P_{norm} \geq Perc_{40}(P_{norm})\ \&\ P_{norm} \leq Perc_{60}(P_{norm})\ else\ \{0$$

$$N_{46} = \{N_{norm}\ if\ N_{norm} \geq Perc_{40}(N_{norm})\ \&\ N_{norm} \leq Perc_{60}(N_{norm})\ else\ \{0$$

Final score is determined using:
Balanced Contrast Similarity Score = $P_{46} \cdot N_{46}$

6. Subheading fonts are selected in decreasing order of balanced contrast similarity score.
7. Results:

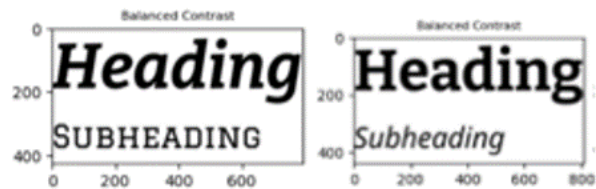

## 4.2 Font pairing algorithm evaluation

For the end-to-end testing of the system we did a controlled A/B test against the current rule based system. The test was conducted against a roll out to 10 percent of our total users, where the divide between **treatment and control was 1:1 ratio**. The roll out was increased gradually to 50 percent. The experiment ran for about 2 weeks before we got a statistically significant result. There was an uptick in the **kept rate of 1.5%** in our designs for the treatment group as compared to the control group. Due to the combination of the A/B experimentation results and the subjective analysis results we decided to go ahead and make this model the default font recommendation system for our generated designs.

## 5. RAI considerations

In order to mitigate any probable RAI issues, each user prompt is parsed through designer's Sev 1 block list and AOAI content safety classifiers to block any objectionable input.

## 6. Conclusions

In this work we created font visual embeddings, font stroke width algorithm, font category to font mapping dataset, a LLM based category utilization description and a lightweight, low latency knowledge distilled mini language model Mini LM V2 model to recommend multiple pairs of contextual heading and subheading fonts for beautiful and intuitive designs. We also utilized a weighted scoring mechanism, nearest neighbor approach and stratified sampling to rank the font pairs as well as bring novelty to the predictions. This model is currently deployed in Design Creator mini app of Designer: **Microsoft Designer - Stunning designs in a flash.** In terms of wider applications and make can be the model of choice for font recommendation across different Microsoft applications such as Microsoft Word, Power Point, Paint etc.

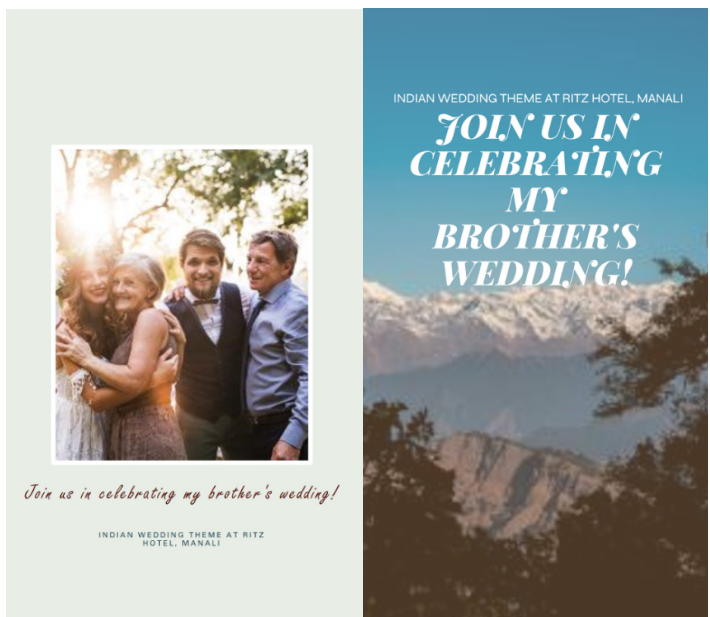

**Figure 5:** Designs powered by font recommendation model.

| Model | BOISSES | SICK-R | STS12 | STS13 | STS14 | STS15 |
|---|---|---|---|---|---|---|
| all-MiniLM-L6-v2 | 81.64 | 77.58 | 72.37 | 80.60 | 75.59 | 85.39 |
| all-MiniLM-L12-v2 | 83.57 | 79.32 | 73.08 | 82.13 | 76.73 | 85.58 |
| e5-base-v2 | 81.40 | 78.30 | 75.79 | 83.58 | 79.95 | 84.46 |
| unsup-simcse-bert-base-uncased | 72.31 | 72.24 | 81.49 | 81.49 | 73.61 | 78.12 |
| GPT Turbo | - | - | - | - | - | - |

| Model | STS16 | STS17 (en-en) | STS22 (en) | STSBenchmark | Subjective evaluation out of 5 | #Parameters |
|---|---|---|---|---|---|---|
| all-MiniLM-L6-v2 | 78.99 | 87.59 | 67.21 | 82.03 | 3.78 | 16M |
| all-MiniLM-L12-v2 | 80.23 | 88.63 | 65.67 | 83.09 | 3.84 | 33M |
| e5-base-v2 | 87.58 | 64.07 | 86.52 | 86.52 | 3.80 | 109M |
| unsup-simcse-bert-base-uncased | 83.58 | 59.65 | 76.52 | 76.52 | 2.13 | 110M |
| GPT Turbo | - | - | - | - | 4.51 | 175 B- |

**Table 1:** Evaluation results for the user prompt to font category retrieval mapping. MiniLM 6-layer V2 model gives comparable performance as compared to much larger models in subjective evaluation at a fraction of latency.